\documentclass[12pt]{amsart}

\usepackage{times}  
\usepackage{helvet} 
\usepackage{courier}  
\usepackage{amsfonts}
\usepackage[hyphens]{url}  
\usepackage{graphicx} 
\usepackage[numbers]{natbib}  
\usepackage{caption} 
\usepackage[a4paper, total={6in, 8in}]{geometry}
\usepackage{hyperref}
\usepackage{xcolor}
\usepackage{blindtext}
\DeclareCaptionType{equ}[][]
\usepackage{float}
\usepackage{tikz}
\usepackage{amsmath}
\usepackage{amssymb}
\usepackage{algorithm}
\usepackage[noend]{algpseudocode}
\usepackage{multicol}
\usepackage{environ}         
\usepackage{etoolbox}        
\usepackage{graphicx}        
\usepackage{soul}
\usepackage{booktabs}

\usepackage[framemethod=tikz]{mdframed}

\newtheorem*{theorem*}{Theorem}   
\newtheorem{prop}{Proposition}[section]

\theoremstyle{definition}

\theoremstyle{remark}

\begin{document}

\title[Honors Thesis, Gaurab Pokharel]{Increasing the Value of Information During Planning in Uncertain Environments}
\author{Gaurab Pokharel}
\address{Department of Computer Science, Oberlin College, Oberlin, OH}
\email{gpokhare@oberlin.edu}

\begin{abstract}
    Prior studies have demonstrated that for many real-world problems, POMDPs can be solved through online algorithms both quickly and with near optimality \cite{Silver_David_Veness,Pineau_Joelle_and_Gordon_Geoff_Thrun_Sebastian, Littman_Cassandra_Kaelbling}. However, on an important set of problems where there is a large time delay between when the agent can gather information and when it needs to use that information, these solutions fail to adequately consider the value of information.  As a result, information gathering actions, even when they are critical in the optimal policy, will be ignored by existing solutions, leading to sub-optimal decisions by the agent.  In this research, we develop a novel solution that rectifies this problem by introducing a new algorithm that improves upon state-of-the-art online planning by better reflecting on the value of actions that gather information. We do this by adding Entropy to the UCB1 heuristic in the POMCP algorithm. We test this solution on the hallway problem. Results indicate that our new algorithm performs significantly better than POMCP. 

\end{abstract}
\maketitle

\section{Introduction}

We as humans instinctively gather information or ask clarifying questions when faced with task completion in uncertain situations. We know to do this because, even though we are delaying the task at hand, it is ultimately in our favour to work with complete information. Ideally, online planning algorithms like POMCP \cite{Silver_David_Veness}, whose sole job is to make plans for agents acting in uncertain situations, know to do the same. They would be able to strategically pick actions that will provide the information to best guide the agent's decision making. However, unlike humans, who can easily correlate information gain with the ease of task accomplishment, these algorithms cannot. It is a difficult task to get them to decide to take any action that explicitly gathers new information.

Moreover, it would be highly unrealistic to assume that machines that employ these algorithms have complete, deterministic information about the environment that they are in. Such domains of problem-solving, where machines do not have deterministic information can be modelled using Partially Observable Markov Decision Processes (POMDPs) and this is what online planners solve. The solution to a POMDP is a sequence of actions called a policy which is the most optimal for an agent to take in order to reach its goal given its current (belief) state. Sometimes, this optimal sequence has one or more sub-sequences where carrying out an action means that the agent needs to collect information. Herein lies the crux of the problem. Existing approaches to solving POMDPs tell the agent that actions that gather information are not actually important because they do not bring the agent physically any closer to the goal. The planning algorithms do not see the \textit{value} in taking this information gathering action. A situation where this happens is when there is a large time delay between when the information is collected and when this information is actually being used. 

Furthermore, real-world applications of these planning algorithms complicate the issue. They come with constraints where the planner has limited time before it needs to gather information, process it, and come up with an action that it thinks is the most viable given the previous two steps. Examples of situations like this can be seen all the time. Almost all computer-assisted human tasks face this problem. When a computer system is assisting a human with a task, to understand what the goal of the user is, the system needs to know well in advance so that it can help the user accomplish the end goal. The computer should be able to ask early in the process or infer what the user is trying to do rather than wait until later even though it will be a long time before ultimately that information might be used. So, how do we get the algorithm to see the \emph{value} in these actions that explicitly gather information, especially when there is a large time delay between when they gather that information and when it's used?

In this paper, we explore a new way in which we can model the planning algorithm in a way that the information gathering actions' value is increased in the planning process. We do this in a computationally inexpensive manner and while preserving the anytime nature of existing solutions. The rest of the paper is organized as follows: Section \ref{Backgrounds} gives the background which is necessary to understand the problem definition detailed in Section \ref{Problems}. Section \ref{Solutions} details the specific solution that this project proposes.  Section \ref{Results} describes the experimental setup designed to evaluate our solution, followed by a discussion of the results of our experiments.  We conclude by summarizing our research and suggesting important avenues of future work in Section \ref{Conclusions}.


\section{Background}
\label{Backgrounds}

\vspace{0.5cm}
\subsection{POMDP Definition} \hfill
\vspace{0.5cm}

A Partially Observable Markov Decision Process (POMDP) is a mathematical framework which allows us to model decision making in an environment where the agent has incomplete and non-deterministic information. Given a belief that an agent holds about the environment, a POMDP framework guides an agent in deciding what action to take, models how the environment might change and what rewards and observations the agent might receive from the environment, and updates the agent's belief based on received observations.  More formally, a POMDP is modeled as a 7-tuple $(S, A, T, R, Z, O, \gamma)$ \cite{Kaelbling_1996} where:

\begin{itemize}
    \item $S$ is the set of all environment states (i.e., the situations an agent can find itself in)
    \item $A$ is the set of all possible actions the agent can take to accomplish tasks and goals
    \item $Z$ is the set of all possible observations the agent can receive that inform its uncertain beliefs about the current state of the environment
    \item $T : S \times A \times  S \to [0,1]$ is the transition function modeling how the environment changes based on actions taken by the agent, where $T(s, a, s') \triangleq Pr(s'|s, a)$ measures the probability of the environment transitioning from state $s$ to $s'$ when the agent takes action $a$
    \item $O : S\times A\times Z\to [0,1]$ is the observation function modeling the information an agent receives for each action it takes, where $O(s', a, z) \triangleq Pr(z|a, s')$ measures the probability of observing $z$ if the agent took action $a$ and the environment transitioned into state $s'$
    \item $R: S\times A \to \mathbb{R}$ details the amount of reward earned by the agent when taking an action in a given state
    \item $\gamma \in [0, 1)$is the discount factor specified by the environment. 
\end{itemize}

\noindent In this research, we assume that $S$, $A$ and $Z$ are finite and discrete, and that each of the above quantities are known \emph{a priori} to the agent.

The next important concept when modeling a problem as a POMDP is \textit{belief}. In partially observable environments, the agent cannot have full information about the current state of the environment. For instance, when planning for a game of chess or go, the agent knows exactly where the pieces are. Versus, a robot on Mars trying to collect good samples where it doesn't know where the samples are or if they are even good.  Instead, in partially observable environment, an agent can only form \emph{beliefs} about what state it \emph{thinks} it is in. A belief $b \in \mathcal{B}$ is mathematically represented as a probability distribution over all possible environment states, where $\mathcal{B}$ represents all possible beliefs. $b(s)$ represents the belief that an agent is in state $s$. These beliefs are calculated and updated using Equation \ref{BeliefUpdate}.

\begin{figure}[H] 
    \begin{equation} 
    \label{BeliefUpdate}
        b'(s') = \eta \cdot O(o \mid s', a) \cdot \sum_{s \in S} T(s' \mid s, a) \cdot b(s)
    \end{equation}
    \begin{small}
        \caption*{\textit{Where $\eta = \frac{1}{Pr(o \mid b, a)}$ is a normalizing constant with $Pr(o \mid b, a) = \sum_{s' \in S} O(o \mid s', a) \sum_{s \in S} T(s' \mid s, a) \cdot b(s)$. $s'$ is the state at time step $t+1$ and $s$ at $t$.}}
    \end{small}
\end{figure}

Since the agent does not know exactly what state it is in, it must calculate expected rewards for taking actions based on the probabilities in its belief $b$:

\begin{equation}
R(b, a) \triangleq \sum_{s \in S}{b(s)R(s, a)} = \mathbb{E}\left[R(s, a) | b\right]
\label{beliefReward}
\end{equation}

The solution to a POMDP model is a policy $\pi: \mathcal{B} \to A$ that prescribes what action the agent should take when it holds beliefs $b \in \mathcal{B}$ in order to maximize the (discounted) sum of rewards it will earn while operating in the environment:

\begin{equation}
V_\pi(b_0) = \mathbb{E} \left[\sum_{t=0}^\infty{\gamma^tR(s_t, a_t)} | \pi, b_0 \right] = \sum_{t=0}^\infty{\gamma^t R(b_t, \pi(b_t))}
\label{expUtility}
\end{equation}

\noindent Here, we try and maximize the \emph{discounted} rewards to take into account the uncertainty in the environment. This is important because the agent does not know exactly what future situations it will find itself in since the state transition function is stochastic, and because it will only receive incomplete information (through observations) about those future states. It receives an immediate reward when it picks an action, but it's future rewards are less predictable due to the inherent uncertainty. We want to take this into account when we are planning and, as such, current rewards get full weight whereas further rewards are weighted less because they are more uncertain. $\gamma$ is the weighting factor\footnote{$\gamma=0.95$ is a common value for POMDPs to give almost full weight to future rewards, but still allow for some discounting} by which we control how much importance we want to give to future rewards.

To estimate the expected discounted sum of rewards (Eq.~\ref{expUtility}) and determine the optimal policy, the agent solves a recurrence relation known as the Bellman Equation:

\begin{equ}[H]
 \begin{equation}
\label{qval}
Q(b,a) =  \sum _{s \in S} b(s) \cdot R(s,a) + \gamma \cdot \sum_{z \in Z} Pr(b' | b, a, z)\cdot V(b')  
\end{equation}
\caption*{\textit{Where $R(s,a)$ is the immediate reward of executing action $a$ in state $s$, $Pr(b'| b, a, z)$ is the probability that the agent ends up in belief $b'$ if the agent executes action $a$ in belief $b$ and receives $z$ as the observation, and $V(b')$ defines the utility of belief $b'$ if the agent continues to follow its current policy. }}
\end{equ}

\noindent  An intuitive interpretation of this equation is that it gives us the \emph{total utility} of picking an action $a$ in belief $b$ at a certain point in time in terms of the \emph{immediate reward} and the \emph{future utility} of the remaining sequence of actions that result from that initial choice.  With this equation, we define two important terms:

\begin{equation}
    \label{V}
    V(b) = \max Q(b, a), a \in A 
\end{equation}

\noindent represents the future utility from a belief $b$ (which creates the recurrence in Eq.~\ref{qval}), and

\begin{equation}
    \label{policy}
    \pi(b) = \text{argmax}_{a \in A} Q(b, a)
\end{equation}

\noindent determines the policy the agent should follow to maximize its discounted sum of rewards (i.e., utility). It is important to note that, the computation of a local policy for any algorithm is entirely dependent on Equations \ref{qval}, \ref{V}, and \ref{policy} and this problem is NP-Hard \cite{NP-Hard}. 

\vspace{0.5cm}
\subsection{Offline vs. Online Planning} \hfill
\vspace{0.5cm}

There are two contrasting approaches to estimating the Bellman Equation and constructing the policy an agent should follow: offline and online planning. 

Offline planning involves finding a \emph{global} policy that prescribes the best possible action that an agent can take for \emph{every} belief $b \in \mathcal{B}$ \textit{before} it operates in the environment.  Thus, the agent goes into the environment with existing knowledge of what it needs to do for any situation.  Popular and/or state-of-the-art algorithms for offline planning include PBVI \cite{PBVI}, HSVI \cite{HSVI}, SARSOP \cite{SARSOP}, Perseus \cite{Percy}.  However, offline planning is really only tractable for problems with small state spaces where only a small number (maybe hundreds or low thousands) of beliefs are actually reachable by the agent from its initial belief.  For larger problems, with thousands, if not millions of states, the problem of finding a policy for every possible belief becomes increasingly complicated and time consuming due to the nature of Equations \ref{qval}, \ref{V} and \ref{policy}. Furthermore, of all the possible belief states in $\mathcal{B}$, not all of them are even reachable, so computing a policy for them would be a waste of precious time and resources. 

In contrast, online planning algorithms generally compute good \emph{local} policies for only the belief that the agent currently holds, then re-plans after it receives an observation and forms a new belief about the environment's state. Thus, the agent interleaves planning and acting as it operates in the environment (see Algorithm \ref{Online_Pseudocode}).  This is generally done through look-ahead search to find the best action to execute at each time step in the environment. Popular and/or state-of-the-art algorithms for online planning include AEMS2, LSEM \cite{LSEM}, POMCP \cite{Silver_David_Veness}, DESPOT \cite{Despot}. These are anytime algorithms which means that they can return a valid solution to a problem even if it is interrupted before it ends. These algorithms are expected to find better solutions the longer they keep running. However, planning in real-time almost always means that there is a time constraint. As such, these anytime algorithms are able to return a policy that the algorithm has so far found to be the best as they calculate approximate solutions, instead of calculating exact values in Equation \ref{qval} for all beliefs. The more time these planners get, the better these approximations get, and the better policy they will generate.

In online approaches, we can generally bound the error of the approximate solution \cite{Ross_OnlinePA}, finding useful policies with fewer computational resources. Furthermore, in extremely large problem domains, offline algorithms become intractable and online planning is the only way to go. As such, in this paper, we mainly concern ourselves with online planning algorithms, more specifically with POMCP.

\vspace{0.5cm}
\subsection{Policy Trees}\hfill
\vspace{0.5cm}

\begin{figure}[h]
    \centering
    \includegraphics[width = .5\textwidth]{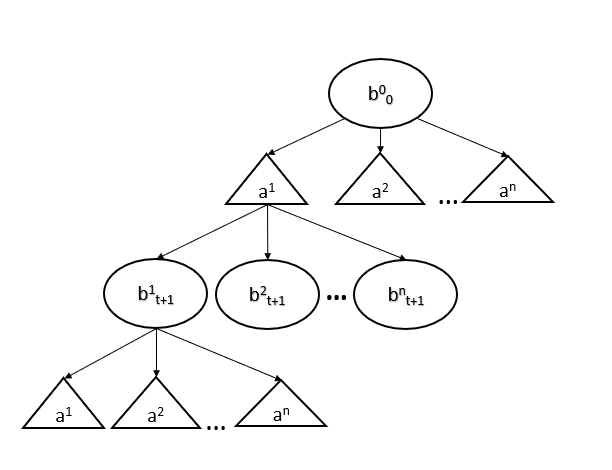}
    \caption{\textit{A policy tree. Each Circular node corresponds to a belief and each triangular node corresponds to an action taken in the belief above it}}
    \label{policy tree}
\end{figure}

One common approach to calculating the Bellman equation (Eq.~\ref{qval}) and generating a policy, especially for online approaches, is to construct \emph{policy trees} consisting of alternating layers of belief nodes and action nodes. The tree is created following a general structure described in Algorithm \ref{Online_Pseudocode}~\cite{Ross_OnlinePA}. The tree (Fig. \ref{policy tree}) starts with an initial belief node corresponding to $b_0$  under which every action that can be performed in the belief are added as action nodes.  For each action, branches are added for all observations that can be produced by the action if it were taken in its parent belief, and the resulting beliefs are added as children nodes along those branches.  As long as there is still planning time and the tree can still grow, the algorithm selects a belief node from the leaves of the tree, expands it by adding its possible actions and their observations underneath the leaf, resulting in new belief node leaves two layers down in the tree.  The Bellman Equation ( Equation \ref{qval}) is performed from the new leaves back up to the root of the tree, and the $Q$ values for each action encountered are updated.

Each node in the policy tree keeps track of important information that is required by a planning algorithm. The action nodes keep track of the $Q$ value for that particular action, given the belief represented by the parent belief node. A belief node keeps tracks of an estimate of the agent's belief at that particular depth (or time step) when it takes a particular sequence of actions and observations before it. To convert the tree into a policy, the action node with the highest $Q$-value determines the action that should be taken if the agent finds itself in the belief represented in the parent belief node.


\begin{algorithm}
\caption{Online Solver for POMDPs \cite{Ross_OnlinePA}}\label{Online_Pseudocode}
\begin{algorithmic}[1]
\Procedure{Online POMDP Solver}{}
\State \textbf{Static:}
\Statex \qquad $b_c$: Current Belief State of agent
\Statex \qquad $T$: Root node of the search tree
\Statex \qquad $D$: Search Horizon
\State $b_c \gets b_0$
\State Initialize $T$ to contain only $b_c$ at the root
\While{\textbf{not} \textsc{Execution Terminated()}}
    \While{\textbf{not} \textsc{Planning Terminated()}}
        \State $b^* \gets $ \textsc{Choose Next Node To Expand()}
        \State \textsc{Expand}$(b^*, D)$
        \State \textsc{Update Ancestors}$(b^*)$
    \EndWhile
    \State Execute best action $a^*$ for $b_c$
    \State Perceive a new observation $z$
    \State $b_c \gets \tau(b_c, a^*, z)$
    \State Update $T$ so that $b_c$ is the new root
\EndWhile
\EndProcedure
\end{algorithmic}
\end{algorithm}

\pagebreak

\vspace{0.5cm}
\subsection{POMCP} \hfill
\vspace{0.5cm}

One of the most computationally expensive parts of POMDP planning is calculating next beliefs (using Equation \ref{BeliefUpdate}, which alone is $O(|S|^2)$ and must be performed every time a next belief is considered in the Bellman equation (Equation ~\ref{qval}). Partially Observable Monte Carlo Planning (POMCP) \cite{Silver_David_Veness} is a very popular and widely used online planning that goes around this issue using Monte Carlo Simulations.

Instead of doing complete belief updates using Equation \ref{BeliefUpdate}, POMCP approximates the probability distribution represented by a belief by using unweighted particle filters, with each particle being a representation of a state and a filter being a collection of such particles. The probability of the agent being in a particular state is represented by the number of particles of each type in a particle filter. These particle filters are stored in belief nodes and as the number of particles used goes to infinity, the particle filter comes to represent an exact belief. 

POMCP constructs a planning tree using Monte Carlo Simulation, following the general structure of Algorithm \ref{Online_Pseudocode}. It makes use of the UCB1 heuristic \cite{UCB1} to pick the next node to expand. During the construction of a policy tree, $b_0$ in Fig. \ref{policy tree} starts off with an initial particle filter consisting of a pre-defined number of particles, sampled evenly from all the possible start states in the environment. The algorithm picks out a particle at random, simulates an action on it, and adds the subsequent belief nodes based on the received observation. This particle is propagated down the tree as more belief and action nodes get added. These nodes keep a copy of the particles that pass through them, gradually building up a filter. One iteration of this particle's movement down the tree is called a \emph{trajectory} down the tree. The trajectory ends when either the particle reaches a terminal state, it runs out of planning depth (reaches the planning horizon), or it reaches a leaf node in the policy tree. In the case of the last, the algorithm is going to do a random simulation of the world until it reaches a terminal state, or again runs out of depth.

As these particles pass through the tree, they also receive some utility (Equation \ref{qval}) from the environment for the actions they pick. The corresponding action node in a trajectory keeps track of the sum of these utilities received by each particle. Then, the overall utility for an action node is calculated as the average reward received by the particles that pass through it. As more trajectories pass through an action node, the utility approximation gets better. As such, POMCP saves us from having to make exact reward and belief calculations and makes use of the limited time that is available to planning in favour of figuring out the optimal action. A complete pseudo-code for POMCP is shown in Algorithm \ref{POMCP}.

\begin{algorithm}[H]
\caption{POMCP \cite{Silver_David_Veness}}
\label{POMCP}
\begin{multicols}{2}
\begin{tiny}

\begin{algorithmic}[1]
\Procedure{Search}{Belief Node $b_{root}$}
    \Repeat 
        \If {$b$ doesn't have particles}
            \State {$s \sim \mathcal{I}$}
        \Else
            \State{$s \sim $ Sample Particle from Particle Filter}
        \EndIf 
    \Call {Simulate}{$s, b_{root}, 0$}
    \Until{Timeout()}
\EndProcedure
\end{algorithmic}

\begin{algorithmic}[1]
\Procedure{Rollout}{$s, b,$ depth}
    \If {$\gamma ^ {depth} < \epsilon $} 
        \State{ \Return 0 }
    \EndIf 
    \State{$a \sim \pi_{rollout}(b, .)$}
    \State{ $(s', o, r) \sim \mathcal{G}(s,a)$}
    \State{$b_{new} \gets $ Belief corresponding to $o$ under $a$}
    \State{\Return $r + \gamma \cdot $\Call{Rollout}{$s', b_{new}, $depth + 1}}
\EndProcedure
\end{algorithmic}

\vfill \columnbreak

\begin{algorithmic}[1]
\Procedure{Simulate}{$s, b,$ depth}
    \If {$\gamma ^ {depth} < \epsilon $} 
        \State{ \Return 0 }
    \EndIf 
    \If {$b$ is leaf node and not terminal}
        \ForAll {$a \in A$}
            \State{Add corresponding action node for $a$ under $b$}
        \EndFor
        \State{\Return \Call{Rollout}{$s, b, $ depth}}
    \EndIf 
    \State{$ a \gets argmax_a V(a) + c \sqrt{\frac{\log{N(b)}}{N(a)}}$}
    \State{$(s', o, r) \sim \mathcal{G}(s,a)$}
    \State{$b_{new} \gets $ Belief corresponding to $o$ under $a$}
    \State{$R \leftarrow r + \gamma \cdot$ \Call{Simulate}{$s'$, $b_{new}$, $depth+1$}}
    \State{Add $s$ to current particle filter}
    \State{$N(b) \leftarrow N(b) + 1$}
    \State{$N(a) \leftarrow N(a) + 1$}
    \State{$V(a) \leftarrow V(a) + \frac{R - V(a)}{N(a)}$} 
    \State {\Return $R$}
\EndProcedure
\end{algorithmic}

\end{tiny}
\end{multicols}
\end{algorithm}

Another important thing POMCP does is making use of the UCB \cite{UCB1} heuristic when picking which action node to expand next (line 8 of the \textproc{Simulate} function in Algorithm \ref{POMCP}). Not all action nodes are equally beneficial to explore for a given belief. Some actions have a higher `potential' for future rewards. POMCP then uses the UCB1 heuristic to introduce an order in which each action node is simulated in the planning process. Paths that are likely to yield a higher reward for the agent are simulated first and get more trajectories. When the planning time runs out, the utility for these good actions are well approximated, consequently always getting recommended as good policy. The heuristic is given by: 

\begin{equ}[H]
\begin{equation}
\label{exex}
    \tilde{Q}(b, a) = \underbrace{Q(b, a)}_{\text{Expected Utility}} + \underbrace{c \cdot \sqrt{\frac{\log(N(b))}{N(b,a)}}}_{\text{ UCB1 Heuristic }}
\end{equation}
\begin{tiny}
\caption*{\textit{$Q$ is the average utility that an action node has accumulated, $N(b)$ is the total number of times that a belief node has been visited, $N(b,a)$ the number of times that this action has been picked in belief $b$ and $c$ is the scaling-factor for the heuristic. At any given phase of the planning process, the action node with the highest value for $\tilde{Q}$ is picked to be simulated. }}
\end{tiny}
\end{equ}

The heuristic is called exploration-exploitation because it is a trade-off between exploiting what the planner already knows about the actions' current utility versus exploring new trajectories for a potentially higher utility. As an action gets simulated many times, the heuristic approaches 0 which motivates the planner to realize that it has the best possible approximation for now and that it should move on to other action nodes to explore. As more trajectories as passed through an action node, the approximation of the q-value also gets better because it becomes increasingly likely that the particles have encountered all possible rewards. 

This works very efficiently when the agent needs to get physically closer to the target. However, when the algorithm cannot see the true `value', i.e. the $Q(b, a)$ in Equation \ref{exex} of actions that gather information, this selective expansion works against itself.

\section{The Problem} 
\label{Problems}

One of the benefits of using a POMDP as the representation of a decision problem is that it explicitly considers uncertainty in the environment. To reduce this uncertainty and achieve high utility, solvers such as POMCP will balance taking actions that gather information to reduce the uncertainty in the agent's beliefs and actions that achieve task-oriented rewards. The solver develops such policies because higher certainty about the true (hidden) state of the environment implies that the agent can, with less error, pick the sequence of actions that gives it the maximum reward. In other words, the probability that an agent takes an action leading to an unexpected bad reward decreases the more certain its knowledge of the current state.

This phenomenon is accurately demonstrated in the Tiger Domain \cite{Kaelbling_1996}. This domain consists of an agent that needs to choose from two identical doors. One of the doors has a tiger behind it and the other a pot of gold. Opening the door with the tiger gives the agent a reward of -100 while opening the other gives provides a reward of +10. The agent starts with a $50\%$ belief that the tiger is behind the left door and a $50\%$ belief that the tiger the right door (representing complete uncertainty about the tiger's location), implying that the expected reward of opening either door is -45. At this point, the agent should choose a third action that represents listening for the tiger, which provides an imperfect (but mostly accurate) observation about the location of the tiger. Say that the agent hears the tiger on the left, and the likelihood of that observation was $80\%$, then the belief update (Equation \ref{BeliefUpdate}) results in a new belief that the probability that the tiger is behind the left door is $80\%$ and $20\%$ behind the right door.  Now, the expected utility of opening the left door is -78 and opening the right door is -12. The expected utility still is not good enough for the agent to decide on which door to pick. The agent should choose to listen again. If it hears the tiger on the right at this time, the belief goes back to $50\%$ (i.e. essentially the root) and the process repeats. But if the agent hears the tiger on the left, it results in a new belief with the probability that the tiger is behind the left being $96\%$. Now, the expected reward for opening the left door is -95.6 and for opening the right is 5.6. Then, the agent picks the door on the right. Here, although the listen action cannot directly give a reward like opening a door, the agent still chooses to gather information before looking for the pot of gold to minimize its risk and maximize its expected rewards in the future.  

We define this benefit of gathering information as the \textbf{value of information ($voi$)}, which can be modelled by quantifying the expected increase in future rewards by performing an information gathering action over the alternative actions:

\begin{equation}
    \label{VOI}
    voi(b,a) = Q(b,a) - max_{a' \in \{A \setminus a \}}Q(b, a')
\end{equation}

\noindent In words, the value of the information gathered from an action at belief $b$ from taking an action $a$ is the difference in the expected reward for the action minus the best-expected reward among all the other actions. If $voi$ is positive, the agent should take an information gathering action (e.g., listening for the tiger); if $voi$ is negative or zero, the agent will pick a task-oriented action (e.g., opening a door). 

For the agent to accurately reason about the $voi$ of information gathering actions, the agent must have a good approximation of the $Q$ value for each action in the current belief $b$.  Given an infinite amount of planning time, all algorithms (including POMCP), will form such good approximations, and the agent will choose information gathering actions when they are most appropriate.  However, when planning time is limited, planning algorithms such as POMCP bias their search for the best policy by spending more effort approximating some $Q$ values over others.  Unfortunately, the actions whose $Q$ receive more effort (i.e., more trajectories during Monte Carlo simulation) are often not those that gather information because gathering information naturally delays when tasks are completed and rewards are received.

To see why this happens, consider a situation where there is a large number of steps between picking an information gathering action $a$ and when the corresponding observation $z$ received is ``cashed in" to receive a positive task reward $r$.  This time delay causes the policy tree to have an exponentially increasing number of possible trajectories between taking action $a$ and receiving the reward  $r$. The planner would have to take a very particular sequence of actions \emph{after gathering information} for it to realize that  $r$ is received as a direct result of earlier observing $z$. For example, when a computer system is assisting a user to accomplish a task, it should very early in the process ask or infer what the user is trying to do, successfully help the user in every relevant sub-task associated with job completion, and receive good feedback from the user. The more sub-tasks there are, the higher the likelihood of the agent not planning for the right sub-task. Planning with POMCP in the Tiger domain does not exhibit this issue because the large reward is at best 2 time-steps away from when the agent chooses to first listen. This means the planner only has to check at most 16 trajectories to find the large $voi$. This means the likelihood of seeing the $voi$ for a listen action is high. This is not exactly the case in larger domains where there are hundreds, if not thousands, of possible trajectories. 

Even if the search for a good policy were \emph{uniformly} biased across all the trajectories between action $a$ and its resulting large reward $r$, there would still be a low probability that the planner will test the right sequence of actions that gives $r$ after information gathering action $a$.  However, heuristics such as $\tilde{Q}$ (Equation ~\ref{exex}) from POMCP bias search towards actions that achieve sooner (albeit possibly lower) rewards since those will be found sooner while expanding the policy tree one node at a time, making it less likely for information gathering action $a$ to be sampled.  This will result in that action experiencing \emph{fewer} trajectories, resulting in both  (1) a lower-quality approximation of $Q(b, a)$ necessary to see a positive $voi$, and (2) fewer child nodes expanded so that the high future reward $r$ that relied on observation $z$ might never even be found.  Consequently, POMCP does not perform well in environments where there are long time delays between when information needs to be gathered and when that information will result in a large future reward. 

We have to make sure that the algorithm is testing these information-gathering actions that actually have a high $voi$ enough times to know that they have a high $voi$. A glaring example of when the $voi$ calculation becomes virtually impossible is the Hallway Problem \cite{Littman_Cassandra_Kaelbling} (a modified version that makes the issue more apparent) is described below. 

\vspace{0.5cm}
\subsection{Long Hallway Domain} \hfill
 \vspace{0.5cm}
 
\begin{figure}[h]
    \centering
    \includegraphics[width = .5\textwidth]{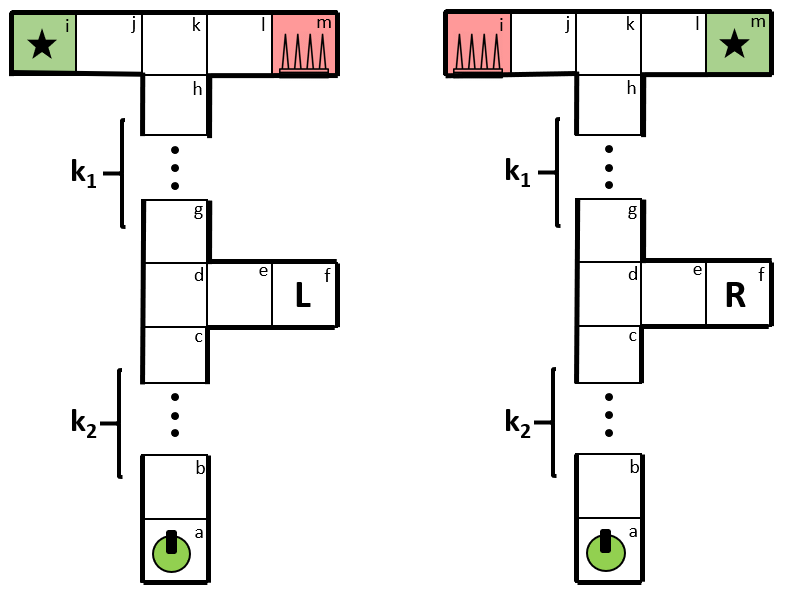}
    \caption{\textit{The Long Hallway problem. $k_2$ determines how long from the start until the agent can gather important information in the small horizontal hallway (where the agent can observe which hallway it is in) and $k_1$ controls the time between when information is collected and when it is used}}
    \label{hallway}
\end{figure}

 \qquad To demonstrate this problem with POMCP, we introduce a novel benchmark domain called the Long Hallway domain, which generalizes a classic, yet easier benchmark called the Hallway problem \cite{Kaelbling_1996}.  This new benchmark is illustrated in Figure \ref{hallway}.
 
 In this new domain, there are $((18 + 2\times k_1 + 2 \times k_2) \times 4)$ states: there are four orientations in each of the $(18 + 2\times k_1 + 2 \times k_2)$ rooms, for instance in figure \ref{hallway} the agent shown is in the state ‘facing north in room a’ state. There are 8 goal states and 8 trap states -- the two stars (denoted by stars) and two traps (denoted by pits) each with the four different orientations, respectively. Then, there are 48 observations that the agent could possibly make, the relative locations of the four different walls, whether the room is an ordinary room without special observation, or the special `left' and `right' observations in rooms f $(2 \times 2 \times 2 \times 2 \times 3 = 48)$. The set of actions consist of wait, move forward, move backward, turn left, and turn right. The agent’s task is to enter the room marked with the star (which provides a reward of +100), and it needs to actively avoid the room with the trap (which provides a penalty of -100 rewards). Taking any other action in the domain results in the agent getting a reward of -1. The movement in this domain is deterministic, implying that the agent can with a 100\% probability execute a selected action given that the resulting state would be valid (i.e. one of the orientations in one of the rooms). The same is true for Observations as well, the agent has perfect observation and can with 100\% probability sense the walls and special rooms.

Existing solutions like POMCP work just fine in bringing the agent physically closer to it’s goal states. However, the same cannot be said for when the planner needs to get the agent to pick the left or right action at room $k$. This is because there are two identical hallways with the star and the trap switched. The agent has no idea as to which of the identical hallways it is in. Because the particle filters are random, the agent may end up with a $50\%-50\%$ probability that it is in either of the two hallways. However, just the difference of $0.1\%$ in those probabilities is enough to make the agent pick an action. Since the agent does not know exactly what state it is in, it must use expected rewards to pick actions. Expected reward of $0.2$ for a belief of $50.1\%$ vs $49.9\%$ is a terrible metric to base a policy where the agent could potentially be collecting +100 or -100. The optimal situation would be where we could get the agent to pick actions that will increase its certainty about where it is in the environment much earlier, multiple steps away from getting a positive or negative reward. 

In the case of the long hallway problem, the optimal strategy is to travel down the small side hallway at $d$ all the way to the end and gather information to know for certain that it is in one of the two hallways. POMCP fails to do this in the case of Hallway because, to the algorithm, it just looks like adding extra steps before it can eventually get to a positive reward state. It cannot reason about the fact that receiving the observation at room $f$ is optimal. This is because the observation is not used for many steps before a positive reward is encountered (at least $7 + k_1$ actions between room f and a star). As a result, POMCP will struggle to calculate the $voi$ for going down the horizontal hallway to gather information, which occurs for two reasons. First, bias towards actions with immediate high utility: since this observation does not give reward, it is seemingly ``useless" to the algorithm and the algorithm finds no reason to go down the side hallway. Second, approximations of $voi$ are further worsened by the biased search, decreasing the likelihood of seeing the large $voi$ of going down the side hallway.

This issue could be solved if we could, somehow, be able to bias search down the path that is likely to have high $voi$ even if the agent does not know that the trajectory has high $voi$ yet.

\section{The Solution}
\label{Solutions}

Every existing POMDP solution has a different way of biasing the best action. POMCP, for instance, uses the UCB1 heuristic which calculates the Q-Value and estimates the upper bound of that Q-value to create an optimistic view of how that action might perform. However, using only this ensures that task oriented actions with high immediate rewards are heavily biased. Any action that collects information used in the future will seem unnecessary and as getting in the way of task completion. We want to convince the algorithm that, even though information gathering actions do not look beneficial in the \emph{near term}, they are actually optimal in the \emph{long term}. We need to be able to do so by preemptively signaling the planning algorithm that a specific trajectory down the tree is desired without having to send a maximum number of trajectories down the path (i.e. without depending on the accuracy of the Q-approximation). 

To come up with a solution, let us go back to Equation \ref{VOI} and analyse how $voi$ changes. In the problem section, we talked about how the POMCP algorithm already implicitly takes into account the value of information when picking an action, as seen in the Tiger domain. However, when the approximation of the Q-values are not accurate enough, this advantage is nullified. The next logical question is, how can we approximate the Q-values for the trajectories without actually having to send down particles to approximate them. The short answer is that we cannot. Instead, we propose a different quantity as a \emph{proxy} for the value of information in such situations: we introduce \textbf{entropy} in the heuristic that is used to pick the best node to expand.

Given a belief, the entropy of the belief is measured by the Equation:

\begin{equ}[!ht]
  \begin{equation}
    H(b) = - \sum_{s\in S} b(s) \cdot \log{b(s)}
    \label{entropy}
  \end{equation}
\end{equ}

\noindent Entropy in a belief directly measures how \emph{uncertain} the agent is about the current state of the environment. The more certain it is, the less the entropy. For instance, in the Tiger domain, the agent picks either the left or right door when it is certain enough that the reward is in one or the other, else it chooses to pick the listen action.  That is, the agent picks an action only when the entropy in its belief state is low enough. Similarly, in a lot of real world problems, the agent has high expected reward when it has low uncertainty (i.e., entropy) about the current state of the environment. This implies that the action which results in the greatest decrease in the overall entropy of the system should often have the highest $voi$. 

We believe that adding the reduction in entropy to the heuristic will bias the planer to pick actions that have high $voi$ without having to depend on the number of trajectories sampled down the particular path.  We propose the modification to the heuristic shown in Equation. \ref{pomcpe_heuristic}.

\begin{equ}[!ht]
\begin{equation}
\label{pomcpe_heuristic}
    \tilde{Q(b, a)} = \underbrace{Q(b,a)}_{\text{Reward}} + \underbrace{c \cdot \sqrt{\frac{\log(N(b))}{N(b,a)}}}_{\text{ UCB1 Heuristic }} +  \underbrace{e \cdot\left( \frac{\Delta H}{\sqrt{\log(N(b,a))}} \right)}_{\text {Entropy Heuristic}}
\end{equation}
 \caption*{\textit{Where $\Delta H$ is the maximum reduction in entropy achieved in the tree under action $a$, and $e$ is the scaling factor.}}
\end{equ}

Here, we pick the \emph{maximum} reduction in entropy under a node because we didn't want our heuristic to be \emph{myopic}, i.e. if the reduction in entropy is more than one time step away from current action, we still wanted to bias towards that reduction. Next, we divide the quantity by $\sqrt{\log(N(b,a))}$ so that as the number of trajectories increases, this heuristic decreases in emphasis (similar to UCB1) so that ultimately the agent is still maximizing the $Q$ values.  That is, when we divide the $\Delta H$, we are making sure that as this corresponding action gets visited many times, yet we are gradually giving more weight to the Q-values instead of the heuristic. This is true as, 

$$N(b,a) \to \infty, \left( \frac{\Delta H}{\sqrt{\log(N(b,a))}} \right) \to 0$$

\noindent We used the $\log(N(b,a))$ instead of $N(b,a)$ (as used by UCB1) in the denominator because experimentation showed that a slower decay of entropy resulted in better performance. 

Another very important consideration we had to make about our solution was the heuristic's time complexity. POMDP solvers are limited in their planning time, so we did not want our solution to add to the time complexity. This new heuristic we proposed requires the algorithm to make repeated entropy calculations for a belief node every time a particle is added. Since we are adding onto the UCB1 heuristic in POMCP, we are still using particles and particles filters to maintain belief states. Then, based on equation \ref{entropy}, this calculation of entropy is $O(|S|^2)$, which would slow down planning, resulting in fewer trajectories in a fixed amount of planning time. To that end, we have come with a $O(1)$ solution to updating the entropy of a particle filter in a belief node at time $t+1$, given the entropy of the same node at time $t$, shown in Proposition \ref{entropyUpdate}. An inductive proof of this $O(1)$ update of entropy is provided in the glossary.

\begin{prop}
\label{entropyUpdate}
Given, at time $t$ the entropy $H_n$, the size of particle filter $n$, and the count of particles in the state that received the new particle $c_x$, the entropy at time $t+1$ can be calculated as: 
\begin{equ}[H]
 \begin{small}
  \begin{equation}
  \begin{split}
     H_{n+1} = & \left({ \frac{-1}{n+1}  }\right) \{c_x\left(\log(c_x+1) - \log(c_x)\right) +  \log(c_x+1) \\ &+ n\cdot \log(n) - (n+1)\cdot\log(n+1)\} + \frac{H_n\cdot(n)}{n+1}
  \end{split}
  \end{equation}
  \end{small}
\end{equ}
\end{prop}

The last thing that we needed to consider in our solution design was the question of when the algorithm could trust the entropy reduction from a particular action. If the beliefs under the action node only had a single particle or two identical particles, then the entropy would often be 0 due to insufficient exploration, instead of actual certainty. This would result in the entropy reduction from the action to seem very large, when in fact the belief node's particle filter is not yet representative of its true belief state. We address this issue by making it so that the algorithm trusts the entropy reduction from choosing an action only if a minimum number of particles has passed through it, say $K_{threshold}$. If the number of particles $\leq K_{threshold}$, then the BackPropagate function in Algorithm \ref{POMCPe} is not called and the entropy heuristic will consider only immediate reduction in entropy.

We implemented this idea into an algorithm (henceforth referred to as POMCPe), the pseudo-code for which is outlined in Algorithm \ref{POMCPe}. Each belief node in the algorithm keeps track of its entropy given its current particle filter. Each action node keeps track of the maximum reduction of entropy (in a single time step) under its tree. This is made possible by the \textproc{BackPropagate($\Delta H$)} function which takes $\Delta H$ up the action nodes until the root, replacing the entropy values in the action nodes if $\Delta H$ is greater than the current value. The \textproc{UpdateEntropy($b,s$)} function in the algorithm makes use of Proposition \ref{entropyUpdate} to update the entropy value in the belief nodes. The only time we make a full entropy calculation is when we

\begin{algorithm}[H]
\caption{POMCPe}
\label{POMCPe}
\begin{tiny}
\begin{algorithmic}[1]

\Procedure{Search}{Belief Node $b_{root}$}
    \Repeat 
        \If {$b$ doesn't have particles}
            \State $s \sim \mathcal{I}$
        \Else
            \State $s \sim $ Sample Particle from Particle Filter
        \EndIf 
        \State \Call{Simulate}{$s, b_{root}, 0$}
    \Until{Timeout()}
\EndProcedure

\Procedure{Simulate}{$s, b,$ depth}
    \If {$\gamma ^ {depth} < \epsilon $} 
        \State \Return 0
    \EndIf 
    \If {$b$ is leaf node and not terminal}
        \ForAll {$a \in A$}
            \State Add corresponding action node for $a$ under $b$
        \EndFor
        \State \Return \Call{Rollout}{$s, b, $ depth}
    \EndIf 
    \State $a \gets \arg\max_a V(a) + c \sqrt{\frac{\log{N(b)}}{N(a)}} +$ \Call{Entropy}{$a$}
    \State $(s', o, r) \sim \mathcal{G}(s,a)$
    \State $b_{new} \gets $ Belief corresponding to $o$ under $a$
    \State $R \leftarrow r + \gamma \cdot$ \Call{Simulate}{$s'$, $b_{new}$, $depth+1$}
    \State Add $s$ to current particle filter
    \State \Call{Update Entropy}{$b$, $s$} 
    \State $N(b) \leftarrow N(b) + 1$
    \State $N(a) \leftarrow N(a) + 1$
    \State $V(a) \leftarrow V(a) + \frac{R - V(a)}{N(a)}$ 
    \State \Return $R$
\EndProcedure

\Procedure{Entropy}{$a$}
    \State $n \gets $ count of total particles through $a$ 
    \State $H_p \gets $ Entropy in the parent belief of $a$
    \State $R \gets 0$ 
    \ForAll{$b$ nodes under $a$} 
        \State $n_i \gets $ size of $b$'s particle filter
        \State $H \gets $ Entropy in $b$'s belief
        \State $R \gets R + \left( \frac{n_i}{n} \right) \cdot H$
    \EndFor
    \State $\Delta H \gets (H_p - H)$ 
    \If{$n \geq K_{threshold}$}
        \State \Call {Back Propagate}{$\Delta H$}
    \EndIf
    \State \Return $\Delta H$ + Max Entropy reduction under node $a$
\EndProcedure

\end{algorithmic}
\end{tiny}
\end{algorithm}


\section{Results and Discussion}
\label{Results}

\vspace{0.5cm}
\subsection{Experimental Setup}\hfill
\vspace{0.5cm}

To determine whether our algorithm POMCPe improves upon POMCP for planning in environments where there is a long time delay between when information is gathered and when it is used, we conducted experiments comparing both algorithms in the Long Hallway problem.  In particular, we ran each algorithm 100 times on randomly chosen starting locations (room a of the left hallway or the right hallway) with $K_1 = K_2 = 1$. We averaged the results from 100 runs so that we weren't gauging the performance based on random chance. To measure the performance of each algorithm, we calculated the average discounted reward and the average cumulative reward for these runs. POMCP only has the $c$ value from UCB1 as its hyper-parameter whereas POMCPe has both the $c$ from UCB1 and the scaling factor $e$ from the Entropy Heuristic as its hyper-parameters. Hence, we performed a grid search for the best hyper-parameters that performed the best for each of the algorithms. The agent was rewarded +100 for reaching to good goal, -100 for reaching the bad goal, and -1 for taking any other action within the environment. 

Once we had the full set of results for  $K_1 = K_2 = 1$, we wanted to test if POMCPe would work for increasing values of the $K$'s. As such, we ran further experiments with$K_1 = K_2 = 2$, the results of which have been summarized in the subsection below. 

\vspace{0.5cm}
\subsection{Comparing POMCP and POMCPe} \hfill
\vspace{0.5cm}

As we can see in Table ~\ref{tab:results1}, POMCPe strongly outperforms POMCP in the domain for these values of $K_1=K_2=1$. The cumulative rewards for the most optimal sequence of actions in this domain would be $44.84$ discounted and $88$ cumulative. So our POMCPe algorithm did not quite find an optimal solution, although it was much closer than POMCP.  The fact that our discounted reward was smaller than optimal but cumulative was so close implies that (1) the agent was reaching the goal state almost every run, but (2) sometimes it was taking more steps than necessary to reach the goal, i.e picking actions like wait, or doing then undoing actions. A look into the experiment logs re-enforces this idea. We would need better hyper-parameter tuning to prevent this from happening. Nonetheless, POMCPe does extremely well compared to POMCP, almost $100\%$ of the time picking to go into the small side hallway, getting the special observation, and end up getting the +100 reward. Whereas POMCP never went down the side hallway, thereby never resolving entropy and always keeping the same uncertain belief with a $50\%-50\%$ probability of being in either hallway, therefore not knowing what to do when it reached the top row. Hence, its cumulative reward is close to 0, which is expected given its purely uncertain belief.

\begin{table}[H]
\begin{tabular}{@{}llll@{}}
\toprule
\textbf{Algorithm} & \textbf{Discounted} & \textbf{Cumulative} & \textbf{Parameters}                                           \\ \midrule
\textbf{POMCP}     & 1.9975              & 4.8500              & \begin{tabular}[c]{@{}l@{}}UCB1 = 50, \\ E = N/A\end{tabular} \\ \midrule
\textbf{POMCPe}    & 28.356              & 82.350              & \begin{tabular}[c]{@{}l@{}}UCB1=100, \\ E = 500\end{tabular}  \\ \bottomrule
\end{tabular}
\caption{\textbf{Rewards for Hallway with $K_1 = K_2 = 1$}}
\label{tab:results1}
\end{table}

Since POMCP was behaving so poorly, we wanted to gauge how bad it really is in approximating $voi$. As such, we modified the start state in the domain to be ``facing west in room e" in either one of the hallways i.e. the agent only has to take a single action of going backwards to resolve all of its uncertainty. The results for this modified experiment is shown in Table \ref{tab:resultsMod}

\begin{table}[H]
\begin{tabular}{@{}llll@{}}
\toprule
\textbf{Algorithm} & \textbf{Discounted} & \textbf{Cumulative} & \textbf{Parameters}                                           \\ \midrule
\textbf{POMCP}     & -25.983             & -36.000             & \begin{tabular}[c]{@{}l@{}}UCB1 = 50, \\ E = N/A\end{tabular} \\ \midrule
\textbf{POMCPe}    & 55.172              & 89.930              & \begin{tabular}[c]{@{}l@{}}UCB1=100, \\ E = 500\end{tabular}  \\ \bottomrule
\end{tabular}
\caption{\textbf{Rewards for Hallway with $K_1 = K_2 = 1$}, modified start state}
\label{tab:resultsMod}
\end{table}

\noindent Here again, we can see that POMCP performed terribly, even when it only had to take a single step to resolve its uncertainty. Whereas, POMCPe  never failed to resolve uncertainty first before moving onto task oriented actions. The tendency of the UCB1 heuristic to prioritize task oriented actions works against itself to an astounding degree. 

\vspace{0.5cm}
\subsection{Evaluating with Larger Hallways} \hfill
\vspace{0.5cm}

Now that we experimentally verified the issues that persist within POMCP, we wanted to see how far we can push POMCPe. So, we increased the values of $K_1$ and $K_2$ to see POMCPe's behavior in increasingly dense tree. The results from this experiment is summarized below.

\begin{table}[H]
\begin{tabular}{llll}
\hline
\textbf{Algorithm} & \textbf{Discounted} & \textbf{Cumulative} & \textbf{Parameters}                                            \\ \hline
\textbf{POMCP}     & 5.5139              & 12.60               & \begin{tabular}[c]{@{}l@{}}UCB1 = 142, \\ E = N/A\end{tabular} \\ \hline
\textbf{POMCPe}    & -0.0866             & 52.66               & \begin{tabular}[c]{@{}l@{}}UCB1=20, \\ E = 500\end{tabular}    \\ \hline
\end{tabular}
\caption{\textbf{Rewards for Hallway with $K_1 = K_2 = 2$}}
\label{tab:resultsMod1}
\end{table}

The cumulative rewards for the most optimal sequence of actions in this domain would be $38.52$ discounted and $86$ cumulative. Looking at the results for the discounted reward, we see that this number is lower than POMCP. Actually, the discounted reward is really terrible. This again is because the agent is staying in the environment much longer than intended, accumulating a lot of -1's in the process, leading to a much lower discounted reward. Due to the very same reason, we see a dip in the cumulative reward as well. However, unsurprisingly, the cumulative reward is still much higher than POMCP, implying that POMCPe was able to direct the agent into the smaller hallway and was able to pick more instances of +100 than POMCP. To be precise, POMCP received a reward of +100 62\% of the time whereas POMCPe managed to do the same 95\% of the time. This indicates an astounding success in being able to direct the agent into the small hallway to receive the special observation even in this larger tree. 

\section{Conclusion and Future Work}
\label{Conclusions}

This project set out to point out and rectify a flaw in POMCP: its inability to prioritize information gathering actions in the planning process when there is a large delay between when the information is gathered and when it is used. We achieved great success in doing so by highlighting POMCP's terrible results in the modified start state within the Long Hallway domain and by coming up with a novel algorithm, POMCPe, that performs much better in similar situations. This algorithm works by considering the reduction in entropy (as a direct result of performing an action) in the planning process and biases search towards trajectories that present a maximum reduction. We tested this novel algorithm against POMCP in the Long Hallway with the modified start states, $K_1=K_2 =1$ and $K_1 = K_2 =2$. In each one of these cases, POMCPe was able to direct the agent towards the special observation in the small side hallway and almost always got the +100 reward whereas POMCP did terribly even in the modified start. However, POMCPe seems to get the agent to stay in the environment much longer than necessary resulting in a lower expected discounted and cumulative reward. 

Future work for this project would involve better hyper-parameter tuning to prevent the agent from staying in the environment too long while using POMCPe. Another reason why this could be happening is that in the planning process, there are multiple trajectories with the same reduction in entropy, except one of them has multiple unnecessary actions in it before ultimately resolving entropy. If the planner encounters this longer trajectory before it encounters the most optimal one, it could end up biasing search towards this longer trajectory, ultimately leading to sub-optimal behaviour and the agent staying in the environment too long. To prevent this from happening, we will experiment with discounted entropy propagation within the tree i.e. the farther away the reduction in entropy is, the less the trajectory will be prioritized. 

We would also make the transition and observation function non-deterministic to see how performance would change in the hallway domain. Furthermore, we will also test POMCPe in domains other than the hallway and see if the results extend to them as well. Once we have more results over multiple domains, we will work on establishing a rigorous theoretical analysis for why POMCPe works and that it performs just as well or better than POMCP. 

This project has effectively demonstrated that POMCPe can increase the value of information gathering actions while planning in uncertain environments.

\section{Acknowledgements}

This project would not have been made possible without the continued support of Dr Adam Eck in the Computer Science Department at Oberlin College and the Oberlin College Computer Science department. 


\bibliographystyle{plain}
\newpage
\section*{Glossary}

\subsection*{Proof that calculation of new entropy is O(1) operation} \hfill
\vspace{0.5cm}

\begin{tiny}
\textbf{Proposition \ref{entropyUpdate} } After adding a new particle to the particle filter that represents an agent's belief state, the entropy $H$ of the particle can be updated as: 

\begin{equation*}
    H_{n+1} = \left({ \frac{-1}{n+1}  }\right) \left[{  c_x\left(\log(c_x+1) - \log(c_x)\right) + \log(c_x+1) + n\cdot \log(n) - (n+1)\cdot\log(n+1)   }\right] + \frac{H_n\cdot(n)}{n+1}
\end{equation*}

\noindent Where, $c_x$ is the count of particles in the state that received the new particle in $H_n$, $H_n$ is the entropy at time $t$ and $H_{n+1}$ is the entropy at time $t+1$. 

\vspace{0.5in}

\noindent \textbf{Proof by Induction: }
\vspace{0.25in}\\
\noindent \textit{Inductive Hypothesis: } Assume that Proposition 1 holds true for all $n+1 < k \in \mathbb{N}$

\vspace{0.25in}

\noindent \textit{Base Case: } Let us use the mathematical definition of entropy to calculate $H_1$ 
\begin{equation*}
    H_1 = -\frac{1}{1}\sum_{s'\in S} p(s').log(p(s')) = \frac{-1}{1}(1\cdot0) = 0 
\end{equation*}
Now, there are two possible cases for $H_2$, let's look at them separately. 

\begin{itemize}
    \item[] \underline{Case 1}: The added second particle is repeated in the filter. If this is the case, using mathematical definition of entropy , 
    \begin{equation*}
        H_2 = - \sum_{s'\in S}p(s') \cdot log(p(s')) = - \frac{1}{1}(0) = 0
    \end{equation*}
    let us now use the proposition to calculate the same value: 
    \begin{align*}
        H_2 &= - \left(\frac{1}{2} \right)\left[ \log(2) - \log(1) + \log(2) + 1\cdot \log(1) - 2 \cdot \log(2) \right] + 0 = 0
    \end{align*}
    
    \item[] 
    
    \item[] \underline{Case 2}: A different particle is added to the filter. Then using the mathematical definition: 
    \begin{equation*}
        H_2 = - \sum_{s'\in S}p(s') \cdot log(p(s')) = \left(\frac{1}{2}\cdot \log\left(\frac{1}{2}\right) +  \frac{1}{2}\cdot \log\left(\frac{1}{2}\right)\right) = \log(2)
    \end{equation*}
    let us now use the proposition to calculate the same value: 
    \begin{align*}
        H_2 &= - \left(\frac{1}{2} \right)[0\cdot (\log(c_x +1) - \log(c_x)) + \log(1) + 1 \cdot \log(1) - 2\cdot \log(2) + 0] = \log(2)
    \end{align*}
    As, we can see equation* 2 and 3 are equal and equation*s 3 and 4 are equal in both the cases. 
    
\end{itemize}
\end{tiny}

\begin{proof}
\begin{tiny}
\begin{align*}
    \begin{split}
        H_{n+1}  &= \left( \frac{-1}{n+1} \right)\bigg[  c_x\left(\log(c_x+1) - \log(c_x)\right) + \log(c_x+1)  \\
                &\qquad + n \cdot\log(n) - (n+1)\cdot\log(n+1) \bigg] + \frac{H_n\cdot(n)}{n+1}
    \end{split}
    \\[2ex]
    \begin{split}
         &= \left( \frac{-1}{n+1} \right)\bigg[ \sum_{s'\in S}c_{n+1}(s')\cdot \log(c_{n+1}(s')) - \sum_{s'\in S}c_{n}(s')\cdot \log(c_{n}(s')) \\
                &\qquad + \sum_{s'\in S}c_{n}(s')\cdot \log(n) -\sum_{s'\in S}c_{n+1}(s')\cdot \log(n+1) \bigg] + \frac{n\cdot H_n - H_n + H_n}{n+1}
    \end{split}
    \\[2ex]
    \begin{split}
         &= \left( \frac{-1}{n+1} \right)\bigg[ \bigg(\sum_{s'\in S}c_{n+1}(s')\cdot \log(c_{n+1}(s')) -\sum_{s'\in S}c_{n+1}(s')\cdot \log(n+1) \bigg) \\
                &\qquad - \bigg(\sum_{s'\in S}c_{n}(s')\cdot \log(c_{n}(s')) - \sum_{s'\in S}c_{n}(s')\cdot \log(n)  \bigg)\bigg] - \frac{H_n}{n+1} + H_n
    \end{split}
    \\[2ex]
    \begin{split}
          &= \left( \frac{-1}{n+1} \right)\bigg[\sum_{s'\in S}c_{n+1}(s')\cdot \bigg( \log(c_{n+1}(s')) -\log(n+1) \bigg) \\
                &\qquad - \sum_{s'\in S}c_{n}(s')\cdot \bigg(\log(c_{n}(s')) - \log(n)  \bigg)\bigg] + \left( \frac{1}{n\cdot(n+1)} \right)\cdot \sum_{s'\in S}c_{n}(s')\cdot \log\left(\frac{c_{n}(s')}{n}\right)  + H_n
    \end{split}
    \\[2ex]
    \begin{split}
         &= \left( \frac{-1}{n\cdot(n+1)}\right)\left[n \cdot \left( \sum_{s'\in S}c_{n+1}(s')\cdot  \log\left( \frac{c_{n+1}(s')}{n+1}\right) - \sum_{s'\in S}c_{n}(s')\cdot \log\left(\frac{c_{n}(s')}{n}\right)\right) - \sum_{s'\in S}c_{n}(s')\cdot \log\left(\frac{c_{n}(s')}{n}\right) \right]\\
        &\qquad + H_n
    \end{split}
    \\[2ex]
    \begin{split}
          &= \left( \frac{-1}{n\cdot(n+1)}\right)\left[n \cdot \left( \sum_{s'\in S}c_{n+1}(s')\cdot  \log\left( \frac{c_{n+1}(s')}{n+1}\right)\right) -(n+1) \cdot \left( \sum_{s'\in S}c_{n}(s')\cdot \log\left(\frac{c_{n}(s')}{n}\right)\right) \right] + H_n
    \end{split}
    \\[2ex]
    \begin{split}
         &= \left( \frac{-1}{n\cdot(n+1)}\right)\left(n \cdot  \sum_{s'\in S}c_{n+1}(s')\cdot  \log\left( \frac{c_{n+1}(s')}{n+1}\right)\right)  + \left( \frac{1}{n\cdot(n+1)}\right)\left((n+1) \cdot  \sum_{s'\in S}c_{n}(s')\cdot \log\left(\frac{c_{n}(s')}{n}\right)\right) + H_n
    \end{split}
    \\[2ex]
    \begin{split}
          &= \left( \frac{-1}{n+1}\right)\left(\sum_{s'\in S}c_{n+1}(s')\cdot  \log\left( \frac{c_{n+1}(s')}{n+1}\right)\right)  + \left( \frac{1}{n}\right)\left(\sum_{s'\in S}c_{n}(s')\cdot \log\left(\frac{c_{n}(s')}{n}\right)\right) + H_n
    \end{split}
\end{align*}

\begin{align*}
    H_{n+1}  &= \left( \frac{-1}{n+1}\right)\left(\sum_{s'\in S}c_{n+1}(s')\cdot  \log\left( \frac{c_{n+1}(s')}{n+1}\right)\right)\\
    &= -\sum_{s'\in S}\frac{c_{n+1}(s')}{n+1}\cdot  \log\left( \frac{c_{n+1}(s')}{n+1}\right)
\end{align*}

\end{tiny}
\end{proof}

\end{document}